# Modeling Shared and Individual Structure for Cross-Subject Continuous Affect Regression from EEG-fNIRS

Xuan Wang, Bing Wang, Shuai Chang, Hao Yuan, Xinbo Qi, Xinyue Zhang

## ABSTRACT

Continuous, second-by-second valence-arousal estimation from physiological signals is typically studied in a subject-dependent setting, where the model sees labeled data from the same person it is later evaluated on. We study the harder zero-shot cross-subject variant on a synchronized EEG-fNIRS dataset: predict raw-scale ([1, 255]) valence and arousal trajectories for subjects whose labels the model never observes, given only their unlabeled EEG/fNIRS recordings while watching the same video stimuli as a disjoint set of training subjects. We decompose the affect trajectory into a structure shared across subjects who watch the same stimuli and an individual structure estimated for each test subject from a label-free EEG marker (alpha-band cross-channel synchrony), which rescales the shared trajectory around the scale midpoint. We validate the per-subject calibration mechanism on four independent axes: leave-one-subject-out correlation between the marker and each subject's true optimal gain, a functional-form comparison against non-linear alternatives, a repeated leave-4-out component ablation isolating each part of the pipeline's contribution, and a ceiling analysis bounding the remaining headroom for per-subject scaling. On held-out subjects, the model reaches an overall MAE of 25.96 / 22.80 across two evaluation batches (valence 21.94 / 19.6, arousal 29.98 / 26.0), well below EEGNet and ASAC-Net baselines reported for the same subject-independent split (raw-scale score 60.6 and 55.0 respectively). We further report a systematic negative-result search across model architectures, feature representations, and prediction targets that found no signal able to improve on the single alpha-synchrony marker.

## 1 Introduction

Continuous affect regression from physiological signals is difficult even in the subject-dependent setting; the harder, cross-subject zero-shot version asks a model to predict valence and arousal for subjects whose labels it never sees, using EEG/fNIRS recorded while they watch the same video stimuli as a disjoint group of labeled training subjects. This setting matters because it determines whether an affective brain-computer interface can be deployed on a new user without a lengthy per-user calibration session — the practical bottleneck repeatedly identified in cross-subject EEG emotion recognition surveys [1]. EEG makes this hard for two structural reasons: non-stationarity of the signal even within a single subject over time, and large inter-subject variability driven by both anatomical differences and individual differences in how strongly felt emotion translates into reported affect amplitude. Because held-out subjects contribute no labels at evaluation time, any per-subject adjustment must be driven entirely by a signal computed from that subject's own unlabeled recording, ruling out personalization techniques that assume access to a handful of labeled target-subject examples [2].

Our model combines a population-average trajectory — the natural baseline given that all subjects watch the same stimuli — with a per-subject gain derived from EEG alpha-band synchrony, in the spirit of label-free test-time calibration methods recently explored for cross-subject neural decoding in other modalities [3]. This report: (1) describes a fully auditable pipeline (Section 4) that substantially outperforms published EEGNet and ASAC-Net baselines on the same subject-independent evaluation protocol; (2) validates the per-subject calibration mechanism directly rather than by assumption, with leave-one-out correlation evidence, a functional-form comparison, and a component ablation isolating the mechanism's own contribution (Section 5); (3) quantifies the remaining headroom for per-subject scaling and reports a systematic, three-axis negative-result survey (architecture, feature representation, prediction target) intended to narrow the search space for future work on this task (Section 5).

## 2 Related Work

The standard assumption in early EEG emotion recognition work — that train and test data come from the same subject or distribution — breaks down once a model must serve a new user with no labeled calibration data. Surveys of this cross-subject generalization problem [1] organize solutions into statistical/adversarial distribution alignment, structural/topological modeling via graph networks, representation learning, generative modeling, and multimodal fusion. Benchmark datasets in this line — DEAP [4], SEED/SEED-IV/SEED-V [5], DREAMER [6], AMIGOS [7] — are dominated by EEG and use largely static, per-clip labels; datasets with continuous, dynamically changing annotations (CASE [8], CEAP-360VR [9]) have so far recorded only peripheral physiological signals rather than direct brain activity. The dataset used here (Section 3) combines both properties — synchronized EEG and fNIRS with continuous, real-time valence-arousal annotation [10]. Personalization methods that report reliable cross-subject gains typically rely on labeled target-subject samples [2], which our zero-label setting rules out; structure-aware test-time calibration methods such as SATTC [3] (cross-subject EEG-to-image retrieval) and T3A-style test-time classifier adjustment [11] are closer in spirit, calibrating a model's output from unlabeled target-subject structure alone. Our per-subject gain calibration (Section 4.2) is an instance of this family, applied to continuous regression and specialized to a single physiologically-motivated scalar. Recent EEG-fNIRS fusion work explores graph convolutions over channel topology and cross-modal attention [12, 13]; our own ablation (Section 5.2) finds that a comparable learnable fusion network does not transfer across subjects in this zero-shot setting.

## 3 Dataset and Task Formulation

We use a synchronized EEG-fNIRS emotion dataset in which 32 participants watched 15 emotion-eliciting video clips spanning five targeted valence-arousal quadrants, with 64-channel EEG (1000 Hz) and 51-channel fNIRS (47.62 Hz) recorded simultaneously [10]. Each participant continuously reported their affective state via a joystick-controlled cursor in 2D valence-arousal space, producing 1 Hz trajectories on the integer scale [1, 255] (neutral point (128, 128)). We adopt a strict subject-independent protocol: 24 of the 32 participants fit every component of the pipeline; the remaining 8 are held out entirely, further split into two disjoint batches of 4 (Test-A, Test-B). The task is continuous regression, $\hat{y}_{1:T} = f(x^{EEG}_{1:Le}, x^{fNIRS}_{1:Lf})$, $y_t = [v_t, a_t]$, scored by mean absolute error (MAE) on the raw [1, 255] scale, averaged across dimensions as the primary metric.

## 4 Method

### 4.1 Population-Mean Trajectory

For every (video, timestep) pair we compute a robust central estimate of the label across the 24 training subjects — a Huber M-estimator followed by a length-5 median filter over time — independently for valence and arousal, giving one trajectory per video, popmean[video, t], shared across all test subjects.

### 4.2 Per-Subject Gain from EEG Alpha-Band Synchrony

We predict a deviation from the scale midpoint and scale it by a per-subject scalar: pred = 128 + gain_subject * (popmean[video, t] − 128). gain_subject is estimated from a label-free marker computed on the test subject's own EEG: the mean absolute off-diagonal cross-channel correlation of alpha-band (8-13 Hz) power across all 64 channels ("sync"), consistent with the established link between alpha-band synchronization and cortical arousal. The deployed gain is a clipped linear function of sync, gain = clip(G * (1 + k * slope * (sync − xm)), b_min, b_max), where slope/xm come from an OLS fit of sync against each training subject's own optimal gain, and G/k are a shrinkage strength and sensitivity multiplier. A safe fallback (gain = 0.90) is used if a subject's recording is too short or the synchrony estimate is non-finite. Figure 1 summarizes the pipeline.

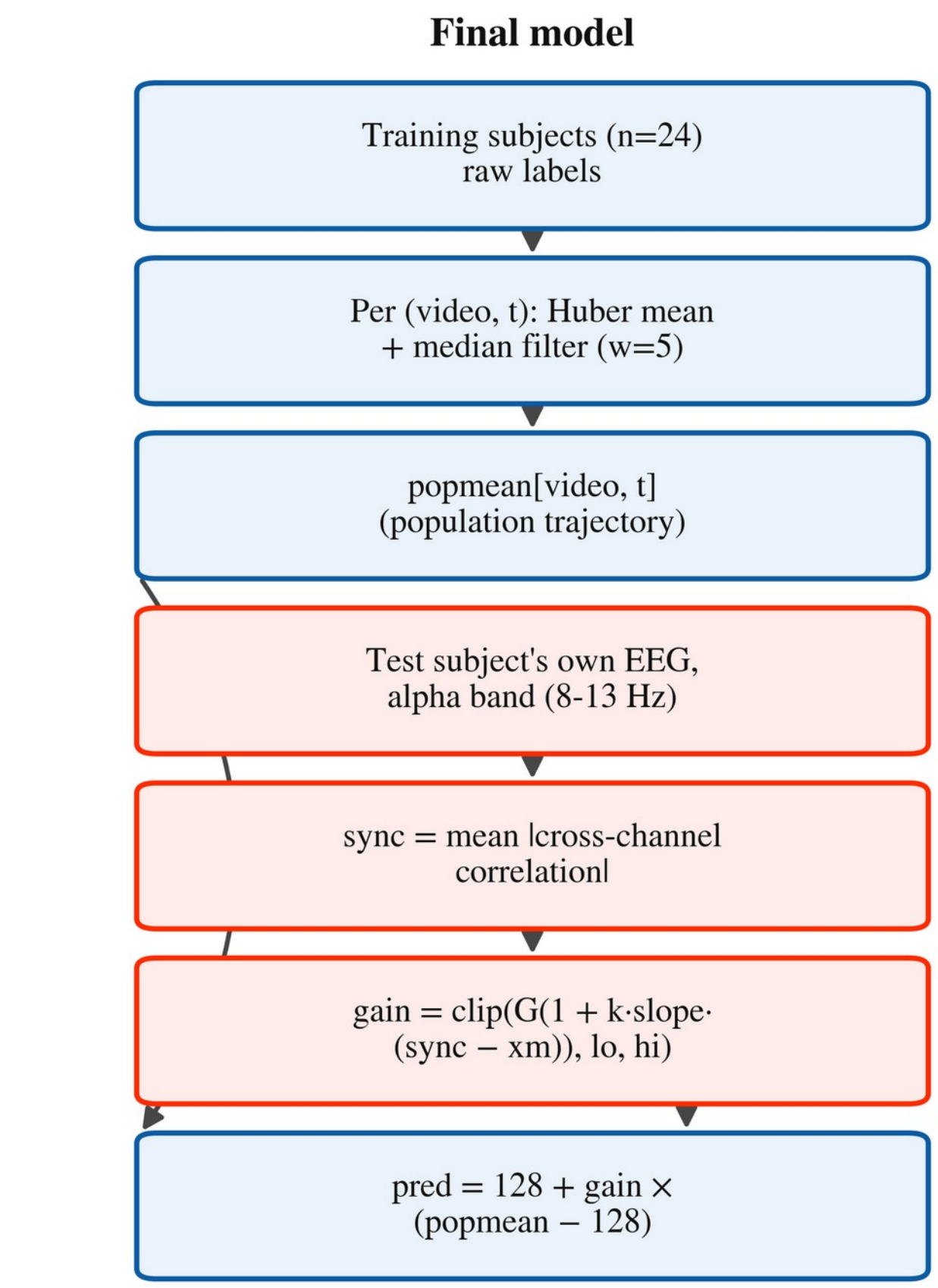

**Figure 1: The model computes a population trajectory from training-subject labels (blue) and a single scalar gain from the test subject's own EEG (red), combined multiplicatively around the scale midpoint.**

# 5 Experiments

## 5.1 Main Results

Table 1 compares our model against EEGNet and ASAC-Net, evaluated on the same subject-independent 24-train/8-test split, both using their published EEG+fNIRS configuration.

**Table 1: Overall test MAE (raw [1, 255] scale).**

| Model | Valence MAE | Arousal MAE | Overall MAE |
|---|---|---|---|
| EEGNet | 63.3 | 57.9 | 60.6 |
| ASAC-Net | 58.6 | 51.3 | 55.0 |
| Ours (Test-A, n=4) | 21.94 | 29.98 | 25.96 |
| Ours (Test-B, n=4) | 19.6 | 26.0 | 22.80 |

## 5.2 Component Ablation

We isolate each component's contribution with a repeated leave-4-out cross-validation on the 24 labeled subjects (N=100 random splits, fixed seed), and separately evaluate the identical four-rung ladder once, directly on the held-out test subjects, as an independent check.

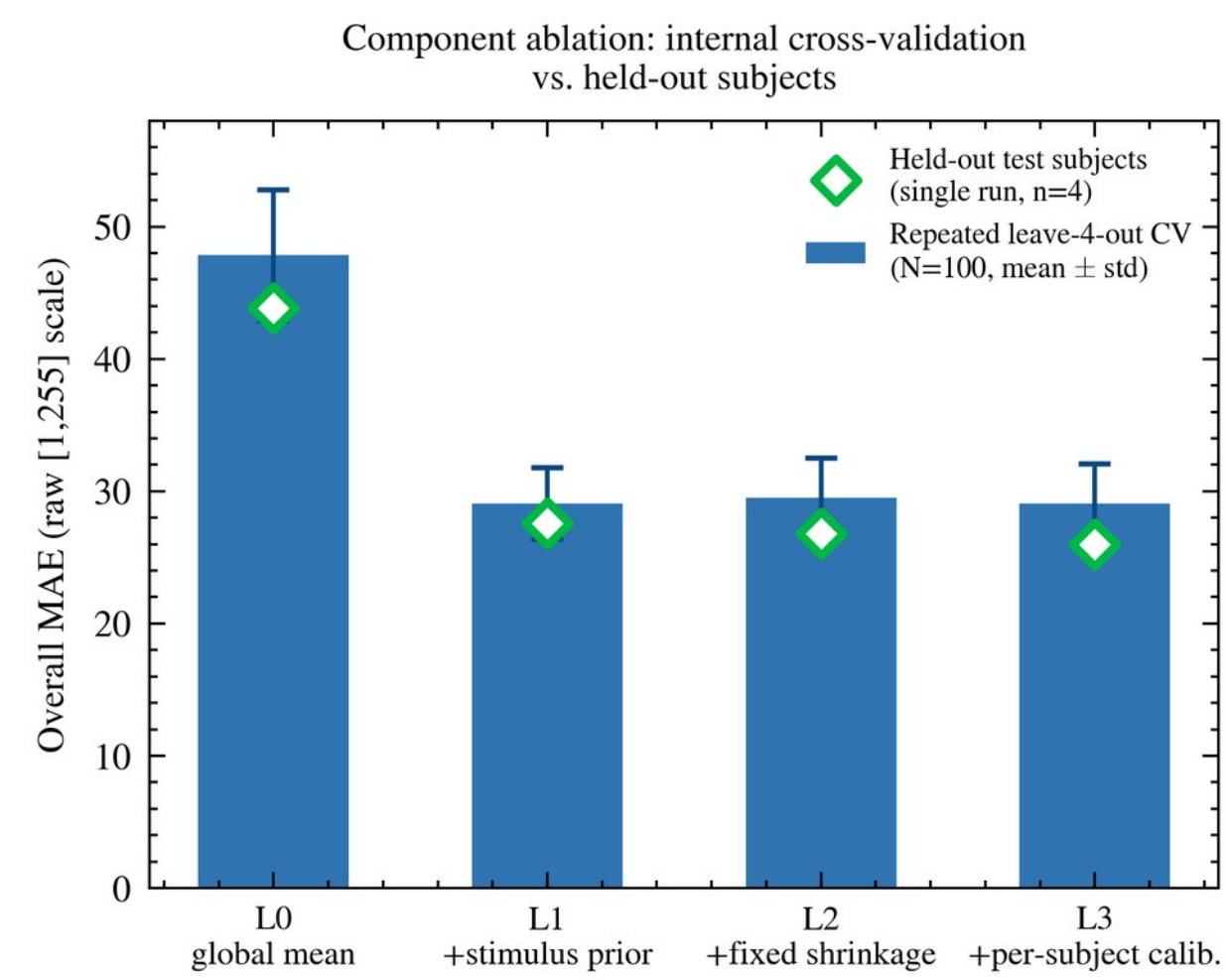

**Figure 2: Component ablation ladder. Bars: repeated leave-4-out cross-validation (N=100, mean ± std). Diamonds: the same ladder evaluated once on held-out test subjects (n=4, no error bar).**

The stimulus-conditioned population trajectory (L0⟶L1) is the dominant component in both series. A fixed, subject-agnostic shrinkage (L1⟶L2) is not distinguishable from noise internally (leave-4-out-optimal shrinkage is G=1.00; G=0.90 costs 0.45 MAE against a fold std of ≈3.0), while the per-subject calibration step (L2⟶L3) produces a small, consistently reproducible improvement (−0.45 MAE, same direction in every configuration checked). The held-out ladder replicates the same qualitative ordering. We separately ablated a learnable multimodal residual network (graph convolution + cross-modal attention + GRU) predicting each subject's deviation from the population trajectory; under the strict subject-disjoint protocol its cross-subject generalization gain is statistically indistinguishable from zero, so it is excluded from the deployed model (λ=0) — a negative result indicating the difficulty here is cross-subject transfer, not model capacity.

### 5.3 Validating the Calibration Mechanism

On the 24 training subjects (n=24), sync correlates with each subject's oracle gain at $r = -0.465$, 95% CI [−0.731, −0.076], $p < .05$ for the shared gain actually deployed, driven mainly by arousal ($r = -0.616$, $p < .01$); valence alone is not significant ($r = -0.324$, 95% CI includes zero), so we do not extend the mechanism to a valence-specific variant. Leave-one-out refitting produces zero sign flips across all 24 subjects. Quadratic, isotonic, and kernel-local alternatives to the linear sync-to-gain form did not improve on it (LOO-MAE 0.194 vs. 0.195-0.210), so we kept the simplest form. A leave-one-subject-out ceiling analysis, comparing the deployed gain against oracles that see the held-out subject's own true labels, shows the deployed calibration is already within 1.44 ± 0.52 MAE (arousal) and 3.01 ± 1.00 MAE (valence) of a perfect scale-only oracle, with a further affine (scale+shift) oracle adding only 0.75-0.84 MAE beyond that — i.e., the remaining headroom for any per-subject scaling mechanism on this input is small.

### 5.4 Negative-Result Search

Because arousal shows the larger gap to its ceiling, we searched for exploitable per-subject arousal signal along three axes: architecture (kNN/mutual-NN trajectory matching, BiLSTM/ridge residual learning, a graph-transformer [14], DTW barycenter averaging, CORAL domain adaptation), feature representation (fNIRS prefrontal markers, phase-amplitude coupling, alpha contrast, multiscale/band-ratio features), and prediction target (direction classification, per-video bias correction). None improved on the single sync scalar (Table 2).

**Table 2: Systematic negative-result search (abridged).**

| Axis | Idea | Verdict |
|---|---|---|
| Architecture | Residual learning (BiLSTM/ridge/LSTM) | Failed (Δ+0.48 to +2.79 MAE) |
| Architecture | Graph-transformer / CORAL | Underperforms; collapses sync variance ~24× |
| Features | fNIRS fusion, PAC, multiscale power | Redundant with sync or no improvement |
| Target | Direction classification | 95.23% acc. = majority-class baseline exactly |
| Target | Per-video bias correction | Failed to improve on baseline |

## 6 Discussion and Limitations

The calibration mechanism collapses each subject's individual differences to a single multiplicative scalar; the ceiling analysis shows this is close to the achievable limit for arousal but leaves more headroom for valence, where the sync marker's own correlation did not reach significance in our sample — closing this gap likely requires a different, valence-specific marker. Our validation of the calibration mechanism relies on 24 training subjects, which limits the precision of both the sync-to-gain regression and the leave-one-out ceiling estimates; a larger subject pool would sharpen both.

## 7 Conclusion

A population-average trajectory rescaled by a single, physiologically-derived per-subject gain substantially outperforms trained EEGNet and ASAC-Net baselines on this subject-independent EEG-fNIRS affect regression task. We validated the calibration mechanism on four independent axes — leave-one-out correlation, functional-form robustness, a component ablation isolating its net contribution, and a ceiling analysis bounding its remaining headroom — and reported a systematic negative-result search across architecture, feature, and target axes that found nothing able to improve on it.